# Vision-Based Robust Lane Detection and Tracking in Challenging Conditions

**Samia Sultana[1], Boshir Ahmed[2], Manoranjan Paul[3], Muhammad Rafiqul Islam[1] and Shamim Ahmad[4]**

[1]School of IT & Engineering, Melbourne Institute of Technology, Sydney, NSW 2000, Australia
[2]Department of Computer Science and Engineering, Rajshahi University of Engineering and Technology, Rajshahi-6204, Bangladesh
[3]School of Computing, Mathematics and Engineering, Charles Sturt University, Bathurst, NSW, Australia
[4]Department of Computer Science & Engineering, University of Rajshahi, Rajshahi-6205, Bangladesh

Corresponding author: Samia Sultana (e-mail: samia.cse.ruet@gmail.com).

**ABSTRACT** Lane marking detection is fundamental for both advanced driving assistance systems and traffic surveillance systems. However, detecting lane is highly challenging when the visibility of a road lane marking is low, obscured or often invisible due to real-life challenging environment and adverse weather. Most of the lane detection methods suffer from four types of challenges: (i) light effects i.e. shadow, glare of light, reflection etc. created by different light sources like streetlamp, tunnel-light, sun, wet road etc.; (ii) Obscured visibility of eroded, blurred, dashed, colored and cracked lane caused by natural disasters and adverse weather (rain, snow etc.); (iii) lane marking occlusion by different objects from surroundings (wiper, vehicles etc.); and (iv) presence of confusing lane like lines inside the lane view e.g., guardrails, pavement marking, road divider etc. In this paper, we proposed a simple, real-time, and robust lane detection and tracking method to detect lane marking considering the abovementioned challenging conditions. In this method, we introduced three key technologies. First, we introduce a comprehensive intensity threshold range (CITR) to improve the performance of the canny operator in detecting different types of lane edges e.g., clear, low intensity, cracked, colored, eroded, or blurred lane edges. Second, we propose a two-step lane verification technique, the angle-based geometric constraint (AGC) and length-based geometric constraint (LGC) followed by Hough Transform, to verify the characteristics of lane marking and to prevent incorrect lane detection. Finally, we propose a novel lane tracking technique, to predict the lane position of next frame by defining a range of horizontal lane position (RHLP) along the x axis which will be updating with respect to the lane position of previous frame. It can keep track of the lane position when either left or right or both lane markings are partially and fully invisible. To evaluate the performance of the proposed method we used the DSDLDE [1] and SLD [2] dataset with $1080 \times 1920$ and $480 \times 720$ resolutions at 24 and 25 frames/sec respectively where the video frames containing different challenging scenarios. Experimental results show that the average detection rate is 97.55%, and the average processing time is 22.33 msec/frame, which outperform the state-of-the-art method.

**INDEX TERMS** Comprehensive intensity threshold range (CITR), ROI, Angle based geometric constraint (AGC), Length based geometric constraint (LGC), Canny edge detector; Lane Detection and Tracking, Novel Lane tracking technique: defining range of horizontal lane position (RHLP); Intelligent Vehicles

## I. INTRODUCTION

The motivation behind the enormous research on advanced driver assistance system, autonomous vehicle, or intelligent transportation system is avoiding vehicle clashes and saving human lives. Lane line detection is the fundamental component of advanced driver assistance systems (ADAS) as lots of traffic rules are based on the lane line mark. Hence, the performance of advanced driver assistance systems depends on lane marking detection to a great extent. Lane detection is the fundamental operation of different ADAS such as the lane departure warning system (LDWS) and the lane keeping assistance system (LKAS) [2, 3]. In the traffic control system, a lane is a part of a roadway that is designed to reduce vehicle conflicts and to guide drivers. Most public roads or highways have at least two lanes separated by lane markings in each direction, which are color yellow, white, or blue. Some automotive companies, such as Mobileye, BMW, and Tesla, etc. have obtained significant achievements by developing some ADAS features.

The prime requirement for any advanced driving assistance system is the system must be real-time because the driver cannot wait for the system to respond till the road accident occurred. Therefore, the lane line detection system must be real-time. In lane detection, it is important to build an accurate and robust lane detection system. For a predefined scenario, accuracy is a systematic closeness to the true values. Robustness is the ability to produce an approximately correct result and to cope with erroneous input and all kinds of challenging conditions. The robustness is the key that determine if a system is appropriate for applying in real life or not. Because of the presence of dramatic variation in the environment of a vehicle roadway, it is very challenging to detect lane marking in real life.

We categorize the significant challenges into four groups that most of the lane detection methods suffer from. Firstly, various change of light conditions affects the visibility of lane marking severely. In a driving roadway, light plays a very important role in lane marking visibility. Thus, different light effects i.e., shadow, glare of light, reflection or low light conditions may affect the lane visibility severely. Glare of light from different light sources e.g., other vehicles headlights or taillights, streetlamp, sun etc. Light reflection from wet road may also interrupt the lane visibility. Additionally, sudden light changes due to change of road surroundings both at daytime and nighttime can severely affect the lane view. For example, when the sun glare appears inside the lane view affect the lane visibility. Again, while passing a tunnel at daytime, sudden darkness appears at tunnel entry. On the other hand, while passing a tunnel or bridge at night, glare of light appears at the tunnel or bridge entry and the tunnel exit area appears as dark out. Extreme or moderate bright colored (white, yellow, blue etc.) light inside the tunnel makes the road shiny and decreases the intensity difference between the lane surface and road surface. Moreover, while lane markings are covered with the shadow of a tree or building, the intensity of the lane marking surface suddenly decreases. Low light conditions caused by rainy or snowy weather also affect the scene. Reflection of any stuff kept inside the car e.g., mobile, navigation device, video or image capturing device etc. on the windshield. Reflection of traffic light, streetlight, vehicle headlight or backlight on the windshield and car hood may also affect the scene. Secondly, depending on the lane type or weather the lane visibility becomes low or obscured. Due to natural disaster like, rain, flood etc. some road markings become eroded and cracked. Again, the visibility of dashed and colored lane is low because the intensity difference between road and lane surface is low. Sometimes, blur visibility created by the frequent presence of raindrops, snow etc. inside the lane view. Thirdly, sometimes lane markings are occluded by wiper, vehicle etc. Occlusion affect the visibility of lane. Finally, misdetection due to presence of lane-like confusing lines e.g., guardrails, pavement marking, road divider, vehicle lines, tree shadow, edges created by cracked road or snow etc.

There is a need to develop a robust lane detection system which can remove the noises creaeted by the light effects along with keeping the lane edges. To eliminate different noises and to ensure successful detection in the subsequent steps video frames are pre-processed. Image preprocessing starts with image smoothing implementing conventional filters such as Gaussian filter [3-5] and Median filter [6]. The priliminary idea of lane detection based on geometric constraint is published in a conference paper [7] where we use Bilateral Filter in a lane detection system for the first time and a comparison between Gaussian and Bilateral filter is presented where Bilateral boost the lane detection rate 10% higher than Gaussian. In this proposed work, we use Bialteral filter to smooth the frames. A comparison among Gaussian, Median and Bilateral filters is shown in figure 4, where Bilateral successfully removed unwanted noises as well as preserving the lane edges.

There are so many edge detectors used to detect the lane edges like Prewitt, Sobel [6], canny etc. [8] compared the performance of Sobel, Prewitt and Canny edge detection techniques and Canny edge detector has the higher accuracy and takes lower execution time than Sobel and Prewitt. In this work, we use Canny edge detector to detect lane edges. Son [9] utilized the Cb component of YUV color space to detect the yellow lane markings. The binary intensity value of a pixel and the Cb value are combined using OR operation. Both the intensity and the yellow value are below the threshold when the yellow lane markings have eroded or occluded. In the proposed work, a comprehensive intensity threshold range (CITR) is introduced in the canny edge detection stage. The pixel intensity value of colored, eroded, blurred, and obscured lane marking is very low. Therefore, to detect the edges of these type of lanes, the intensity threshold is selected 30 and the lower intensity threshold value is selected one third of upper threshold that is 10.

Selecting region of interest (ROI) is a simple and efficient way to reduce computational time and surrounding noise. It is noticeable that the lane line information lies in the lower half of the image. So, we do not need the whole image to process for lane detection. Previous research works select the ROI as the bottom side of the image [10], while others use the vanishing point detection technique [11], [9] to define the ROI. Estimating a vanishing point can be helpful in detecting lanes, because parallel lines converge on the vanishing point in a projected 2-D image. In [6], they selected the region of interest dividing the image scene into road region and sky region by using an adaptive horizon line. This region of interest selection keeps lots of confusing non-lane lines inside region of interest which lead to misdetection. In our work, we select a simple and predetermined region of interest using an isosceles trapezoid-shaped mask placed in middle of the horizontal axis and two third of vertical axis excluding the lower third to avoid the car hood edges. Isosceles trapezoid is trapezoid in which the base angles are equal and therefore the left and right-side lengths are also equal. This mask efficiently keeps the lane line edges while removing partially or fully other confusing noisy edges like, guardrails, pavement marking, road divider, car hood edges etc. This reduces computational time and misdetection as well. Inside a region of interest, the lines are detected by Hough transform which is a feature extraction method for detecting shapes such as circles, lines, etc. by applying a voting procedure. It is the most applied technique to detect lane lines[5, 9, 12]. We applied Hough Transform to detect the position of all the lines inside the ROI. These lines are called candidate lane lines (CLL).

Jung et al. [13] detected lanes based on the spatiotemporal image. They successfully detected lanes with sharp curvature, lane changes, night roads, obstacles, and lens flare due to using the temporal consistency of lane width on each scanline. However, since the detection result of Jung's method depends on the lane width, the system fails to detect the lane while the width of the lane is either increasing or decreasing. Additionally, the spatiotemporal image depends on vehicles' speed, which will affect the detection results. When both the left and right lane is completely missing, this method can't detect lane. In the proposed method, the lane detection and tracking algorithm is not dependent on either width of the lane or speed of the vehicle and able to detect lane while both lane marking is completely missing. Here, we have separated CLL into two groups of lines: candidate left lane lines (CLLL) and candidate right lane lines (CRLL) using slope-based constraint. Left and right lane markings will be detected separately from these two different sets of lines after verification. Therefore, if the lane width varies from road to road, it does not affect the accuracy of the proposed lane detection method.

[14, 15] both detected lane markings based on spatio-temporal incremental clustering and curve fitting.[13] also used cubic model to detect the curved lanes. In our method, we don't need any specialized model or method to detect curved lane. Our method works for both curved lane and straight lane with the same detection and tracking method. In front of a car the closer part of a curved lane appeared as straight line. A frame with curved lane as input and the output of proposed method is shown in figure 1 where only closer part is detected which is enough for any lane detection applications. Figure 2(g) is also an example of successful detection of a sharp curved lane marking.

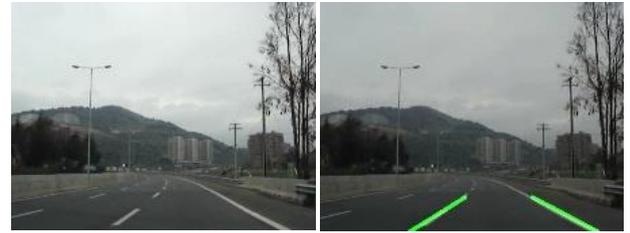

Figure 1: input and output result of a curved road from SLD dataset [2]

Inside the region of interest there is high possibility of misdetection due to existence of objects similar to lane marking. For example, guardrails, pavement marking, road divider, vehicle lines, tree shadow, edges created by cracked road etc. To remove the incorrect line segments [16] determines the angular range for left lane (25,75) and the angular range for right lane (105,155). However, their method cannot remove the lines those are parallel to lane marking. In the proposed method, a two-step lane verification technique is introduced to avoid misdetection: angle based geometric constraint (AGC) and length based geometric constraint (LGC). In the angle based geometric constraint, two range of angle that are determined for left and right lanes are [45°-c, 45°+c] and [135°-c, 135°+c] respectively where c is a variable for the position of camera. As the range of angle defined in our method is narrower than [16], our method can eliminate more confusing lines. After implementing AGC we get filtered candidate left lane lines (FCLLL) and filtered candidate right lane lines (FCRLL). However, there is still possibility of existence of non-lane lines which are either parallel to lane marking or form angle within the angle range. Therefore, in the proposed method, length based geometric constraint (LGC) is introduced as the next step of verification. Usually, the camera is mounted on the middle of the car and the left and right lane is located at the immediate front of the car. Additionally, trapezoid shaped mask also shortens the length of other unwanted noisy lines. Therefore, the line with max length among FCLLL is considered as the left lane marking and the line with max length among FCRLL is considered as the right lane making.

In the preliminary work [17] we propose a lane tracking approach by angle (between left and right lane) validation technique consisting two consecutive phases: initialization and validation. However, it did not work properly while the road is curved, and the position of the car is not centered. The most widely used tracking methods are Kalman Filter [1, 12, 18] and Particle Filter [19]. [6] used Kalman filter to track lane markings and they showed that the time consumption of lane tracking is 2.36 ms per frame. In our proposed method, a novel and simple lane tracking technique is introduced which takes only 0.99ms to process per frame.

It is noticeable from the video frame sequences that; the vertical position of a lane remains almost same but the horizontal position of a lane changes remarkably. Considering this issue, in this paper, we propose a novel and simple lane tracking technique to predict the lane position of next frame by defining a range of horizontal lane position (RHLP) of next frame with respect to horizontal lane position of previous frame. The upper range is created by adding the estimated deviation with previous frame's horizontal lane position and the lower range is created by subtracting the estimated deviation from previous frame's horizontal lane position. It consumes only 0.99ms/frame to track the lane marking.

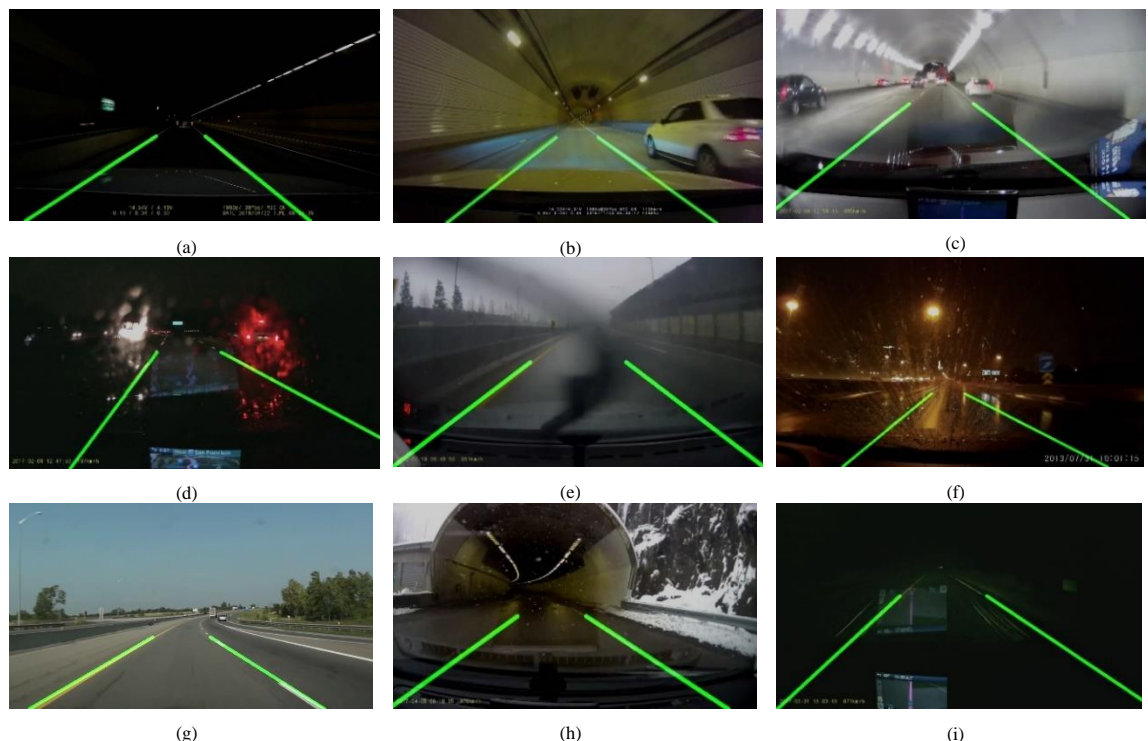

**Figure 2.** Example of Successful lane marking detections by the proposed method under harsh driving conditions from DSDLDE [1]; (a) daytime dark tunnel, (b) nighttime shiny yellow light tunnel, (c) nighttime shiny white light tunnel, (d) inside and outside reflection of light on car windshield, (e) blur view due to heavy rain, (f) reflected streetlight on wet road in rainy night, (g) sharp curve (h) heavy snow tunnel entry, (i) lane like edges created by nighttime snow

In summary, the main contributions of the proposed method are:
- We propose a comprehensive intensity threshold range (CITR) in the edge detection stage, which improves the performance of canny edge detector to enhance and detect the edges of obscured and low-intensity (either colored, eroded or blurred) lane markings.
- An isosceles trapezoid-shaped region is selected as region of interest (ROI) to remove unwanted noisy edges. To detect the lines inside ROI Hough Transform is used. These detected lines are called candidate lane lines (CLL). To develop a lane detection system where left and right lane is not dependent on each other's position, CLL are separated into two groups using slope-based constraint: candidate left lane lines (CLLL) and candidate right lane lines (CRLL).
- A two-step geometric constraint-based lane verification technique is introduced to verify the characteristics of lane and to avoid misdetection:
  i. Angle based geometric constraint (AGC): It is noticeable that left lanes create an acute angle and right lanes create an obtuse angle with x axis correspondingly. Based on this concept, angle based geometric constraint is defined. A range of acute angle is defined for left lane using the middle value of acute angle and a range of obtuse angle is defined for right lane using the middle value of obtuse angle. Therefore, the angle constraint for left lane is [45°-c, 45°+c] and the angle constraint for right lane is [135°-c, 135°+c] where c is the variable for position of camera. Applying AGC we get filtered candidate left lane lines (FCLLL) and filtered candidate right lane lines (FCRLL).
  ii. Length based geometric constraint (LGC): Usually, the camera is mounted on the middle of the car and the left and right lane is located at the immediate front position of the car. Therefore, the line with maximum length among the FCLLL is considered as the left lane marking and the line with maximum length among the FCRLL is considered as the right lane marking.
- We propose a novel and simple lane tracking technique to predict the lane position of next frame by defining a range of horizontal lane position (RHLP) of next frame with respect to horizontal lane position of previous frame. The upper range is created by adding the estimated deviation with previous frame's horizontal lane position and the lower range is created by subtracting the estimated deviation from previous frame's horizontal lane position. This range will be updated automatically according to the lane position of previous frame. This lane tracking technique needs no priori information or any initialization phase to track the lane. It can keep track of the lane position when either left or right or both lane markings are partially or fully invisible due to erosion or occlusion.

The proposed method can detect and track the lane marking under different harsh driving conditions i.e., daytime dark tunnel entry, shiny yellow light tunnel, shiny white light tunnel, glare of taillight from another vehicle, blur view due to heavy rain at both daytime and nighttime, heavy snow tunnel entry at daytime, heavy snow at night etc. as shown in figure 2.

The rest of this paper is organized as follows. Section two reviews research work on lane detection and tracking systems presented in the literature. In section three, we introduce and discuss our proposed lane detection and tracking method. In section four, the obtained experimental results and comparison with other's methods has been presented. Section five concludes the paper.

## II. RELATED WORK

In this section, we discuss the work related to lane marking detection and tracking. A large volume of research has been done on lane marking detection. Most of the lane detection methods are computer vision based. [20] figure out vanishing points using stereo cameras to generate a minimum cost map, while Yoo et al. [21] used geometric relationships between line segments and the vanishing point for lane detection. [22] propose a model to detect lane, based on LiDAR and Around View Monitor (AVM) camera fusion using color filtering for lane markings extraction. [23] used the HSV colour transformation to extract the white features and add preliminary edge feature detection in the preprocessing stage. In [24], they proposed an all-weather lane detection method based on image classification and hybrid isometric operator performed on simulation interactive platform. However, the system can not predict the lane extension curvature and they did not address snowy weather conditions. [25] proposed a system using bird's eye view image with a 2D Gabor filter to enhance lane marking followed by a marking extraction and a Bezier curve fitting technique. In, [9], they overcome illumination change effect by detecting vanishing point based on voting map, defining an adaptive ROI and detecting lanes using invariance properties of lane colors.

In [26], the algorithm estimated the vanishing point fast and accurately and the estimated vanishing point determined the line segments that belong to the lane marking. In [27], authors proposed a gradient-enhancing conversion method that provides strong edges to the lane line in illumination conditions. However, it does not work well in extreme conditions because they assume that one scene does not have multiple challenges. [28] reduced the computational complexity by detecting vanishing points and establishing an adaptive region of interest (ROI) and able to detect lanes under illumination. However, the processing time is 200 msec per frame which is too high to be appropriate for real time application. They addressed only three types of road scenario, night scene, the rainy scene and fluctuating illumination scene. Whereas, proposed method addressed more challenging conditions like rainy day, rainy night, snowy day, snowy night, tunnel etc. [1] processes a gradient cue and a color cue together and a line clustering with scan-line tests to verify the characteristics of the lane markings. They also extract the line segments using Edge Drawing lines (EDlines) method and cluster those lanes according their features. The line segments are tracked using Kalman filter. They added the U-V values and grayscale intensity to boost the values of yellow lane marking. However, this method fails to detect yellow lane markings while the lane markings become obscured due to lighting effects.

Machine or deep learning-based techniques can improve the accuracy of lane detection to a great extend. However, the algorithms have higher hardware requirements, and the training models are too complex [29-31]. [5, 32] detected lane marking using the RANSAC spline fitting technique. [33] introduce improved RANdom SAmple Consensus(RANSAC) algorithm by using the feedback from lane edge angles and the curvature of lane history to prevent false lane detection. However, the computational complexity of RANSAC model is high and their lane detection method takes total 667 ms per frame to detect lane which does not meet the real time requirements. Additionally, these methods suffer from overfitting problem. They usually require lots of data for training to achieve high accuracy. These mehtods are not explainable or interpretable and highly dependent on training dataset. Therefore, they are not robust for all different challenging environment. Whereas proposed method need no training phase and can meet the real time requirements.

## III. PROPOSED METHOD

In this section we discuss our proposed novel, simple and robust lane detection, and tracking method.

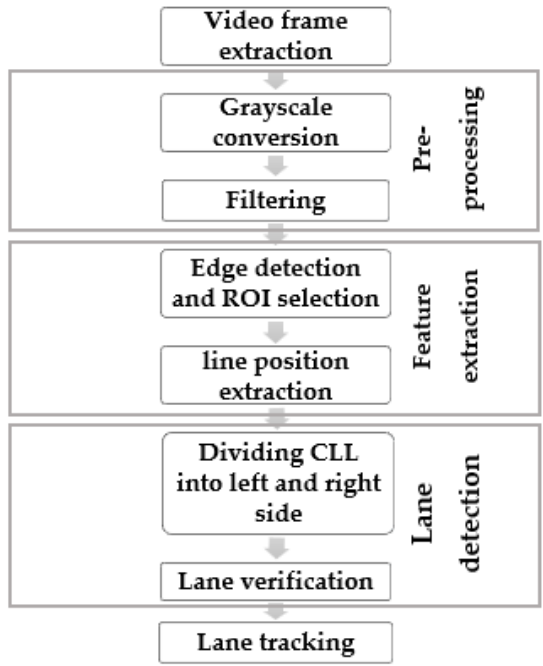

**FIGURE 3:** Block diagram of the proposed lane detection and tracking method

Here, we divided the proposed lane detection and tracking method into four major parts- i) pre-processing, ii) feature extraction and iii) lane detection and iv) lane tracking. The overview of the proposed lane detection method is shown in Figure 3. In the pre-processing stage, the extracted video frames were gray-scaled, and the *bilateral filter* is used to remove noises as well as preserve the edges. Next, In the feature extraction stage, edges were detected by the canny edge detector improved by CITR and the position of lines was extracted by Hough transform. After that, in the lane detection stage, a robust lane verification technique, angle and length based geometric constraint is proposed which successfully verifies candidate lane lines. Finally, a robust lane tracking technique, horizontally adjustable lane repositioning range proposed which predicts the lane location of the present frame using the information of lane location of the previous frame.

### A. PRE-PROCESSING

Pre-processing is an important part of the lane line detection procedure. Sometimes, images contain noises due to challenging environmental conditions: low light, overexposed light, wet road, eroded or colored lane or poor weather conditions (rain, snow, etc.). Noises significantly affect the visibility of lanes as well as the performance of image processing. Therefore, pre-processing is used to enhance the feature of interest and reduce noise. In addition, smoothing is part of pre-processing techniques intended for removing noises without losing image information. We have pre-processed input video frames by gray-scale conversion [34] and noise filtering.

Smoothing or noise filtering is the simplest way to denoise an image. To carry out smoothing operation, Gaussian [10], mean or median [35] filters are used. In [36], to preserve the feature of interest and to remove unwanted clutter the image was filtered by median filter and to enhance the grayscale image, image histogram has been used. To remove different lighting effects, at preprocessing stage adaptive threshold is performed. Adaptive thresholding is performed using Otsu's algorithm. It is observed that Bilateral filter improved the detection rate 10% compared to gaussian filter [7]. Therefore, in this paper, we used bilateral filter to smooth the image.

Bilateral filter is a nonlinear, edge preserving and noise reducing image filter. There is no need to employ any edge sharpening technique after smoothing because bilateral itself preserves the strong edges besides smoothing. The bilateral filter is the weighted average of neighborhood pixels, which is same as Gaussian convolution. The difference is that the bilateral filter considers the difference in intensity value with the neighbors to preserve edges while smoothing. For a pixel to influence another pixel, it should not only occupy a nearby position but also have a close intensity value to that pixel [37, 38]. In an input image $I_p$ is the coordinate of centered pixel, q is the coordinate of the current pixel to be filtered, $I_p$ and $I_q$ is the intensity value of pixels p and q respectively. The equation of bilateral filter, $I_{BF}$ can be expressed as follows,

$$I_{BF} = \frac{1}{W_p}\sum_{q\in S} G_{\sigma_s}(\|p - q\|) G_{\sigma_r}(|I_p - I_q|) I_q \quad (1)$$

In Figure 4, the performance of Bilateral filter has been compared to gaussian and median filter. The input images are smoothed by three different filters to remove noise edges. However, to compare the performance of these filters, canny edge detector is applied on the filtered image to detect both noise edges and lane edges inside the smoothed images. In Figure 4(a) the input image contains flare of light and the reflection of light on wet road which severely affected the visibility of right-side lane marking. In Figure 4(b) the input image includes flare of headlight and taillight from cars, the presence of rain drops on the windshield and the reflection of a navigation device on windshield. These issues severely affected the visibility of both left and right lane markings. Figures 4(c), 4(d) are the output images of the gaussian filter and figures 4(e) and 4(f) are the output of the median filter. Both filters failed to remove the noises mentioned above. In Fig. 4(g) and 4(h) it is clearly visible that bilateral filter removes most of the noises along with keeping the lane edges.

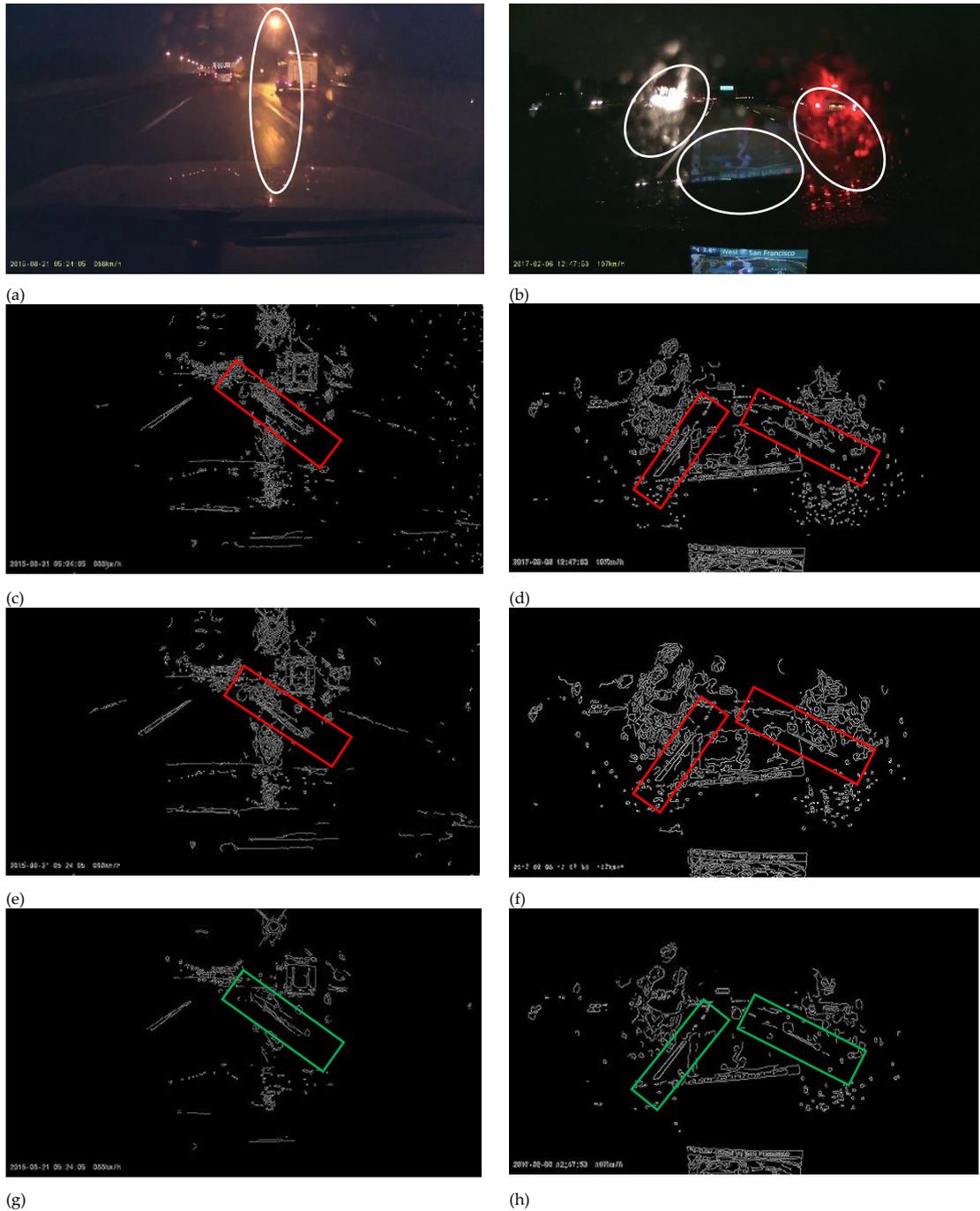

(a)  (b)

(c)  (d)

(e)  (f)

(g)  (h)

**FIGURE 4.** Comparison among Gaussian, Median and Bilateral filters, (a), (b) input images, white ellipses indicate the source of noises, (c), (d) Gaussian filtered binary edge images, (e), (f) median filtered binary edge images, red rectangles indicate the noisy lane edge area, (g), (h) bilateral filtered binary edge images, green rectangles indicate almost clear lane edge area

## B. FEATURE EXTRACTION

At feature extraction stage, particular lane features such as edge, texture, length, width, or color etc. are identified. In the case, when the illumination conditions drastically change and the view becomes blur and obscured specially in the rainy or snowy weather, it is tough to discriminate the road and the lane by using color or texture feature. Edge-based feature is more robust than color-based features in various illumination condition and adverse weather condition. Therefore, in this method, we considered the edge feature for lane detection method.

### 1) EDGE DETECTION AND ROI SELECTION

There are lots of edge detection techniques like Sobel, Canny, Prewitt, Roberts etc. [35] applied a fuzzy method for lane detection and canny to get a better edge detection. In [38], the performance of canny and Sobel has been compared and experiment shows that the canny is better than Sobel. Canny is a multi-step algorithm. At first, noise is removed by gaussian. Then, the edge gradient and direction are determined. Next, an edge thinning technique named non-maximum suppression has been applied. Finally, the candidate edges are detected and connected by dual-threshold method.

This dual threshold method uses two thresholds $[T_u, T_l]$ where $T_u$ denotes upper intensity threshold value and $T_l$ denotes lower intensity threshold value, to find the edges of interest. The edge pixels above the upper limit are accepted as edges and edge pixels below the threshold are rejected. Pixels in-between upper and lower threshold are considered only if they are connected to pixels of upper threshold. The ratio between the upper and lower threshold is recommended as 3:1.

However, it is challenging to choose the exact intensity threshold range for different varying lighting conditions. Specially it is difficult to detect the edges of colored lane (yellow, blue etc.) line due to the low intensity difference between road surface and lane marking. These colored lanes become almost invisible due to blur view while it's raining or snowing. Therefore, we proposed a comprehensive intensity threshold range which improves the performance of *canny operator* to detect lane edges of colored, obscured, and blurred lane marking due to varying lighting conditions e.g., day, night, rainy, snowy etc. As our proposed intensity threshold range can detect all kind of lane edges under different lighting conditions, it is named as comprehensive intensity threshold range (CITR). We have selected the upper intensity threshold value, $T_u$ = 30 which is quite low because in low light or blur visibility condition the intensity of lane edge pixels becomes low. We have chosen the lower intensity threshold value, one third of upper threshold that is $T_l$=10.

In Fig 5(a) there are two input frames where each scene has following multiple challenges: i) the view is blur due to rainy weather, ii) the left lanes are yellow and iii) wipers obscure the lane view. In figure 5(b) and 5(c) the intensity threshold ranges that have been applied correspondingly [50,10] and [45,15]. In both cases the lane edges detected partially. In figure 5(d) the intensity threshold range [30,10] is applied and in this case most of the lane edges detected.

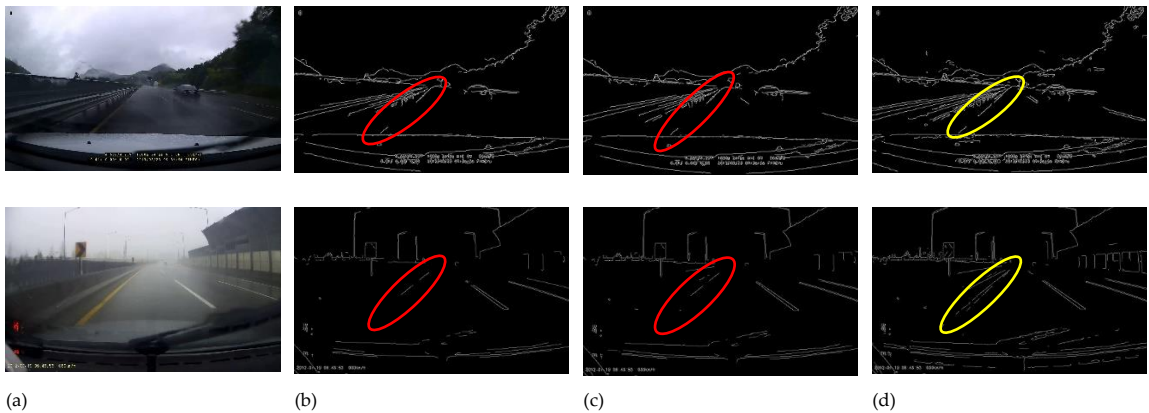

(a)      (b)      (c)      (d)

**FIGURE 5.** Comparison among three different intensity threshold range to detect yellow colored lane edges, (a) input images, (b) Implementation of intensity threshold range, [50,10], (c) Implementation of intensity threshold range [45, 15], red ellipses indicate the failure of edge detection, (d) Implementation of intensity threshold range [30,10], yellow ellipses successfully detected most of the yellow-colored lane edges

We experimentally observed that this is the most comprehensive intensity threshold range for all colored lane lines and for all lighting conditions as it can handle multiple challenges mentioned above. However, CITR will keep lots of noisy edge pixels along with lane edge pixels. Selecting region of interest can resolve this issue.

It is noticeable that the lane line information lies in the lower half of the image. So, we do not need the whole image to process for lane detection. Most of the videos inside datasets are captured by positioning the camera in a middle place behind the windshield. Road is always located in front of the vehicle because vehicle moves in a forward direction [35]. Therefore, it is recommended to select ROI at the lower side of image [10, 35, 36]. We have selected a region of interest using an isosceles trapezoid-shaped mask placed in middle of the horizontal axis and two third of vertical axis excluding the lower third to avoid the car hood edges. Isosceles trapezoid is trapezoid in which the base angles are equal and therefore the left and right-side lengths are also equal. This mask efficiently keeps the lane line edges while removing other noisy edges for different datasets. In Figure 4(d) the output image of region of interest selection has been shown.

### 2) LINE POSITION EXTRACTION

Hough transform is a feature extraction method for detecting shapes such as circles, lines, etc. in an image by applying a voting procedure[6]. In our proposed method, this algorithm is used to identify the position of the lines created by the edges. Two points of each line (x1, y1) and (x2, y2) are the final output of the Hough Transform. Thus, position of all the lines inside the region of interest are being extracted by Hough Transform. Since among these extracted lines only one left and one right line would be the lane lines, these lines are called candidate lane (CL) lines. In figure 6(a), three sample frames has been shown as input. In figure 6(b), 6(c), 6(d) and 6(e) all the corresponding output images for bilateral filtering, canny edge detection, region of interest selection and Hough Transform has been shown respectively.

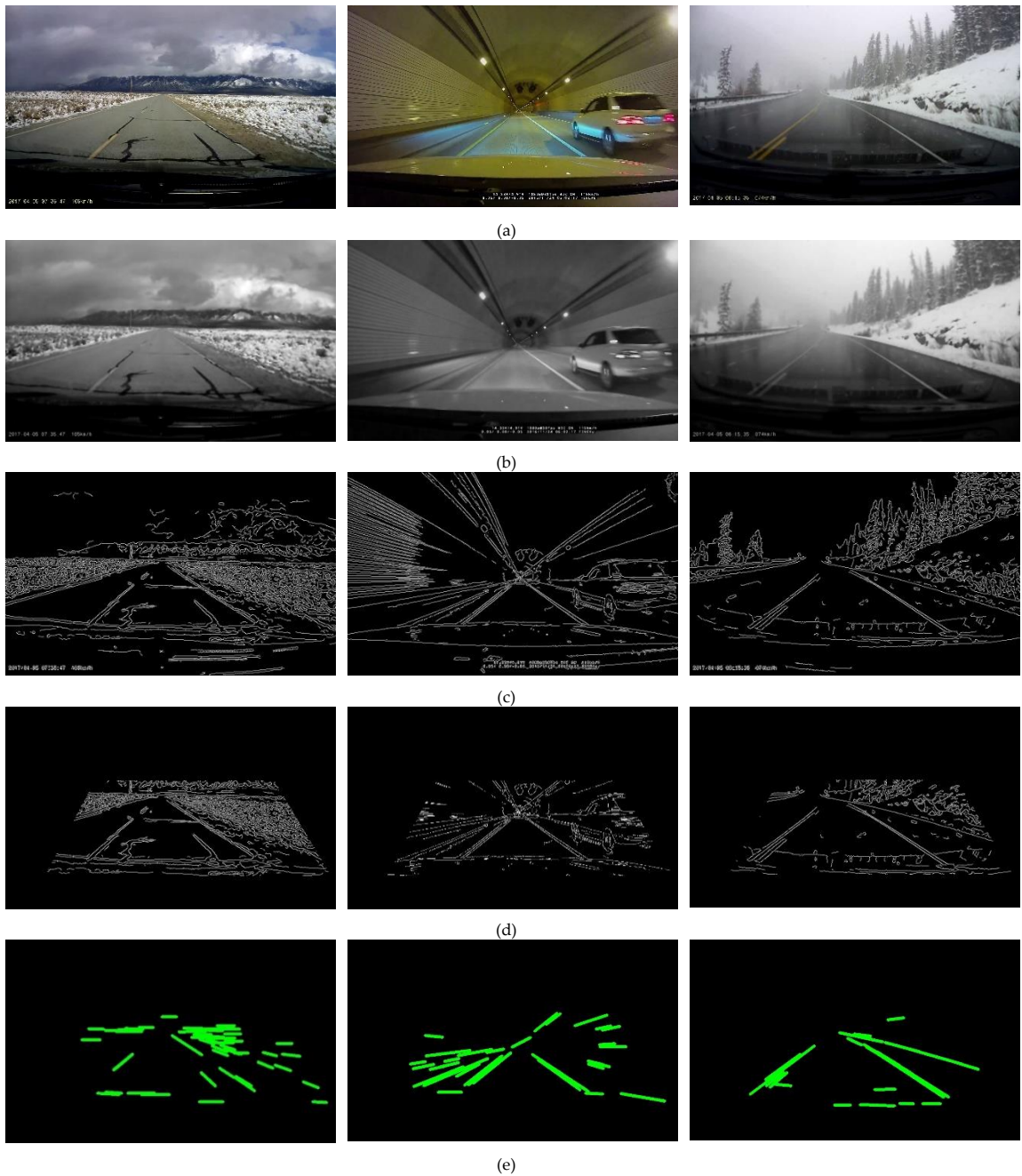

**FIGURE 6.** (a) input frames containing lane marking view affected by cracked road, shiny yellow light effect and snow-covered blurry effect (from left to right); (b) bilateral filtering; (c) CITR based canny edge detection (d) region of interest selection by isosceles trapezoid-shaped mask; (e) lines detection by Hough transform

## C. LANE DETECTION

To detect the left lane and right lane boundary simultaneously in each frame is called lane detection. In this paper, lane detection task is divided into following steps: a) dividing candidate lane lines (CLL) into left and right-side lines, b) lane verification.

### 1) DIVIDING CLL INTO LEFT AND RIGHT SIDE

Hough transform successfully detected all the lines inside region of interest. From figure 5(e) we can see that Hough Transform has detected lots of lines including lane lines. All these lines are called candidate lane (CLL) lines as among them lane lines would be detected. Here we have separated candidate left lane lines (CLLL) from candidate right lane lines (CRLL) lines using slope-based constraint. Slope, m is the ratio of vertical change to horizontal change of a line. A line has a positive slope if y increases along with x and on the other hand slope is negative if y decreases along with x increases. The left lane line always has a negative slope and right lane line always has a positive slope. By using equation (2) the slope is calculated. The slope-based constraint has been set by equation (3). So, lines with slope m > 0 identified as CRL lines, slope, m < 0 as CLL lines and slope m = 0 & m = ∞ as false positive.

$$\text{slope}, \quad m = \frac{\Delta y}{\Delta x} = \frac{y2-y1}{x2-x1} \quad (2)$$

$$Lane = \begin{cases} Right; & m > 0 \\ Left; & m < 0 \\ False; & m = 0 \end{cases} \quad (3)$$

2) Lane Verification

Sometimes false lane detection happens due to presence of other lookalike lines e.g., guardrails, pavement markings, road divider, traffic signs, zebra crossing, car hood, reflection of stuff (mobile, navigation device, video or image capturing device) kept inside the car, vehicle lines, the shadow of trees etc. which have line like structure. Therefore, lane verification is important to perform to filter out these confusing lines. It is noticeable that, from frame to frame, the vertical position of lane lines almost remains unchanged, but the horizontal position (along x-axis) of lane lines gets changed. So, we have set a range of angle formed by lane lines with x axis to verify the horizontal position.

An angle which measures less than 90° is called acute angle. The left lane forms an acute angle(anti-clockwise) with x axis. An angle that measures greater than 90° and less than 180° is called obtuse angle. The right lane forms an obtuse angle(anti-clockwise) with x axis. Therefore, a range of angle is defined for both left and right lane considering these two types of angles. So, the angle between the left lane and x-axis(anti-clockwise) is defined as $\theta_l$ and the angle between the right lane and x-axis (anti-clockwise) is defined as $\theta_r$. According to the definition of acute and obtuse angle, the range of $\theta_l$ should be less than 90 degree and the range of $\theta_r$ should be greater than 90 degree and less than 180 degrees.

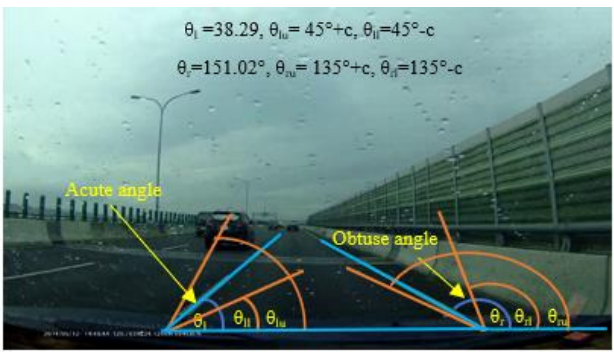

**FIGURE 7.** Angle based geometric constraint, $\theta_{lu}$ and $\theta_{ll}$ is the upper and lower range value of left lane angle $\theta_l$ correspondingly and $\theta_{ru}$ and $\theta_{rl}$ is the upper and lower range value of right lane angle, $\theta_r$ correspondingly

In case of left side lane, to minimize the range and to maximize the noisy line removal, we have selected the mid value between 0 to 90° i.e., 45° as borderline. We considered a variable c for the position of camera. So, the lower value is defined by subtracting c from 45° and the upper value is defined by adding c with 45°. Finally, the range of angle ($\theta_l$) of left lane formed with x axis is $[\frac{\pi}{4} - c, \frac{\pi}{4} + c]$. On the other hand, for right side lane, we have selected the mid value of an obtuse angle range i.e., 135° as borderline. So, the lower value is defined by subtracting c from 135° and the upper value is defined by adding c with 135°. Finally, the range of angle ($\theta_r$) of right lane formed with x axis, is $[\frac{3\pi}{4} - c, \frac{3\pi}{4} + c]$. We experimentally observed thousands of frames with a variety of challenging situations and found that above consideration of angle range performs best. A visual representation of the above proposed angle verification has been shown in Figure 7. We choose a random frame from the

$$\theta = \tan^{-1} m \quad (4)$$

$$angle = \begin{cases} \theta_l, & \frac{\pi}{4} - c \leq \theta_l \leq \frac{\pi}{4} + c \\ \theta_r, & \frac{3\pi}{4} - c \leq \theta_r \leq \frac{3\pi}{4} + c \end{cases} \quad (5)$$

dataset to show the angel ranges. In [16], they have determined the angular range for left lane (25,75) and the angular range for right lane (105,155) which is wider than our range. As range of angle defined in our method is narrower than their method. Therefore, our method can eliminate more confusing lines than their method. All the angles, $\theta$ is calculated by the equation (4) and the mathematical representation of angle range has been shown in equation (5).

Applying angle-based constraints we get filtered candidate left lane lines (FCLLL) and filtered candidate right lane lines (FCRLL). However, this angle-based constraint can't remove some confusing lines which form angles with x axis within the angle range or which are parallel to lane lines. For example, guard rails, road divider, road curb, bridge railing, wall edges of road tunnel, other vehicles, shadow of tree or building, pavement marking parallel to lane etc., these objects create confusing lines because of their angle and position. To avoid this type of misdetection we have proposed length-based geometric constraint.

Usually, the camera is mounted on the middle of the car and the left and right lane is located at the immediate front of the car. So, the lane line is visible as the longest line among all other confusing lines. In some cases, the lane line may not be the longest line, but due to the trapezoid shaped ROI, part of other long lines such as lines of guard rail, road divider, road curb, bridge railing, road tunnel wall etc. is eliminated. In the case of dashed lane line, Hough transform turned the dashed short lines into one long line. Thus, we choose length parameter for final verification. We took certain number of frames and measured the length of different lines created by lane lines, guard rail, vehicle edge, parallel other lane, road crack, tunnel wall edges etc. which form angle with x axis within the angle range and calculated the average length of different type of lines.

In figure 8, shows a comparison between the length of lane marking line and other lines e.g., guard rails, vehicle edges, road crack edges, tunnel wall edges, arrow pavement, wiper etc. The metrics indicates that length of lane line is longer than any other lines. We observed that more than 95 percent cases lane lines are longer than any other lines in each frame. The longest left line is the true left lane line, and the longest right line is the true right lane line. Using equation 6, the length of each FCLL line has been calculated and using equation 7, the length of each FCRL line has been calculated. The line with maximum length among FCLL lines has been selected as left lane line (equation 8) and The line with maximum length among FCRL lines has been selected as right lane line (equation 9).

$$length_{left}(FCLL) = \sqrt{[(x_{l1} - x_{l2})^2 + (y_{l1} - y_{l2})^2]} \quad (6)$$

$$length_{right}(FCRL) = \sqrt{[(x_{r1} - x_{r2})^2 + (y_{r1} - y_{r2})^2]} \quad (7)$$

$$lane_{left} = \max(length_{left}) \quad (8)$$

$$lane_{right} = \max(length_{right}) \quad (9)$$

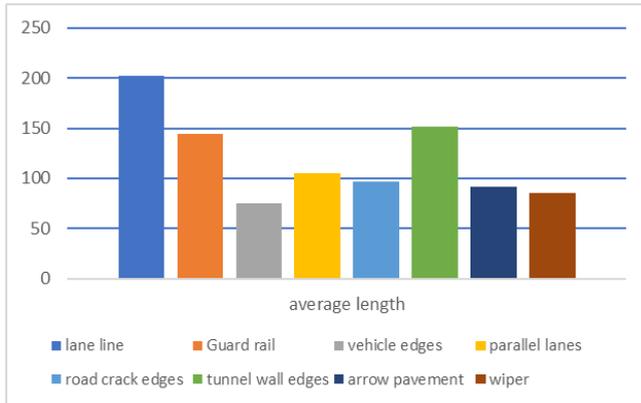

**FIGURE 8.** Comparing length of lane lines with other confusing lines e.g., Guardrail, vehicle edges, parallel lanes, road crack edges, tunnel wall edges, arrow pavement, wiper etc.

### D. LANE TRACKING

To estimate and predict the position of the lane markings of the next frame using the lane marking position of previous frame is called lane tracking. Lane tracking increases the probability to detect lane markings in challenging conditions. Lane tracking is implemented to follow the change of lane position. Sometimes challenging conditions such as rain, snow, reflection of road lamp on wet road, overexposed sunlight, or shiny tunnel light wear out the lane marks and the presence of raindrop on windshield and snow on the road affect the visibility of road marking severely.

Therefore, it is very important to develop a lane tracking system that can detect lane marking even when they are partially or fully invisible for short period of time. For lane tracking, Kalman filter [2, 39], extended Kalman filter [40], Annealed particle filter [41], and super-particle filter [42] are used.

Between two consecutive video frames, there will not be much deviation as there is temporal and spatial continuity between frame sequences. It is noticeable from the video frame sequences that; the vertical position of a lane remains almost same but the horizontal position of a lane changes remarkably. Considering these issues, we propose a novel and simple lane tracking technique to predict the lane position of next frame by defining a range of horizontal lane position (RHLP) of next frame with respect to horizontal lane position of previous frame. The upper range is created by adding the estimated deviation with previous frame's horizontal lane position and the lower range is created by subtracting the estimated deviation from previous frame's horizontal lane position. This range will be updated automatically according to the lane position of previous frame.

A lane line position P can be easily defined by two points $(x_1, y_1)$ and $(x_2, y_2)$. If the lane line position of previous frame is $P_p(x_{p1}, y_{p1}, x_{p2}, y_{p2})$ and the lane line position of next frame is $P_n(x_{n1}, y_{n1}, x_{n2}, y_{n2})$. Since, the position of lane does not change vertically, $y_{p1}=y_{n1}$ and $y_{p2}=y_{n2}$. Therefore, the $P_n$ would be located either at left side of $P_p$ or at right side of $P_p$. We measured the deviation of lane position $P_p$ along x axis (both for left and right-side lane) in more than 300 frames and the average deviation of lane position along with x-axis is less than 6 percent of the width of the image. If the deviation of the lane position is d, the range of $x_{n1}$ is R1, the range of $x_{n2}$ is R2, w is the width of the image, the expression of d, R1, R2 is as follows:

$$d = w * z/100; \; z<6 \quad (10)$$

$$R1 = [(x_{p1} - d), (x_{p1} + d)] \quad (11)$$

$$R2 = [(x_{p2} - d), (x_{p2} + d)] \quad (12)$$

This technique can effectively increase the lane detection rate. After employing lane tracking the system can detect lane even when the car is changing the lane or in the presence of confusing lines that are parallel to the lane lines or while the lane line is fully or partially invisible or unpainted for short period of time. This procedure takes only 0.99ms per frame whereas Kalman filter takes 2.36ms per frame [6]. We have shown how the lane tracking system can adjust while changing lane in figure 8. Here, a lane changing scene has shown by picking six frames from a rainy-day video from frame no. 515 to 677.

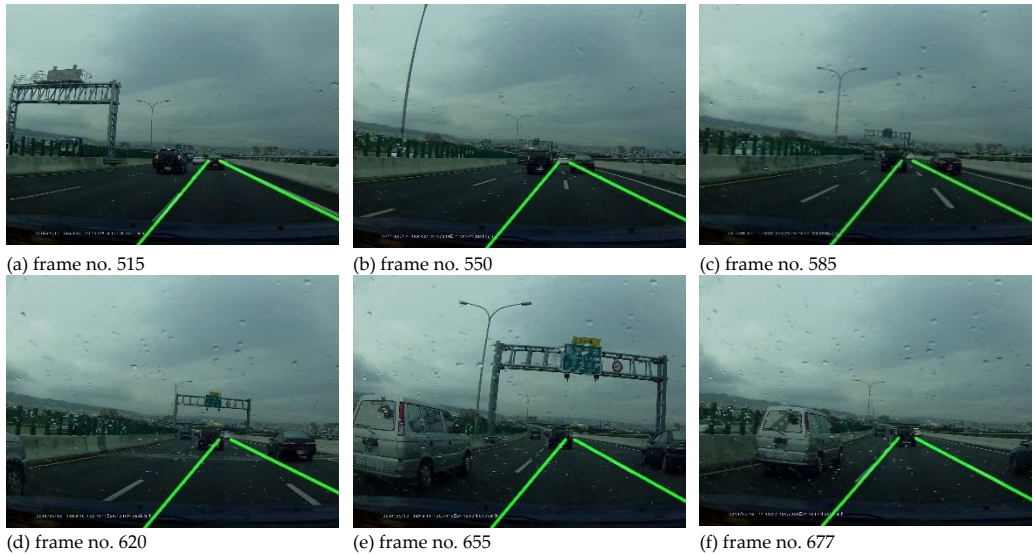

**FIGURE 9.** Automatic lane tracking while lane changing

From figure 9(a) to 9(e) the lane change happens, and the tracking system can keep track of the lane. After shifting to the left lane completely, the tracking method can automatically adjust the lane line in the figure 9(f).

In figure 10 some challenging conditions are shown where lane markings are partially or fully eroded, dashed, unpainted or invisible due to some difficulties like, occlusion by wiper, blurred view due to heavy rain, darker road view without streetlamp, overexposed headlight from opposite direction car, shiny effect and flare of light created by streetlamp reflection on the windshield. All these difficulties were handled by our proposed lane tracking method.

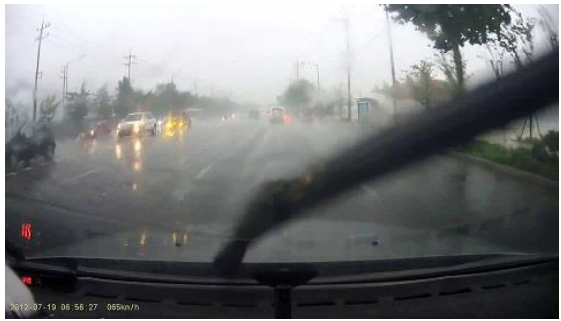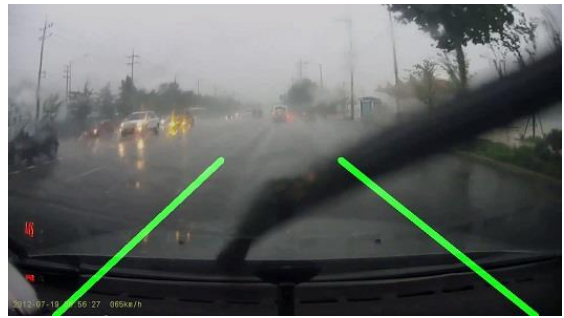

(a)

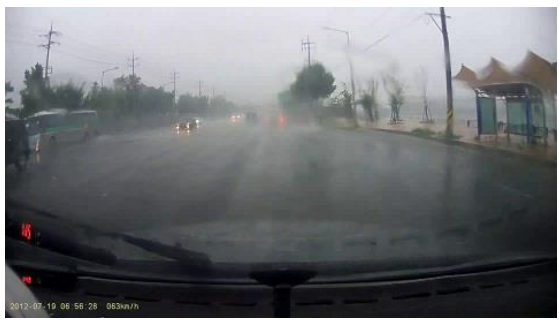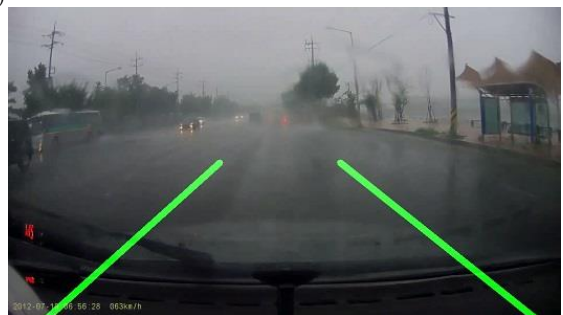

(b)

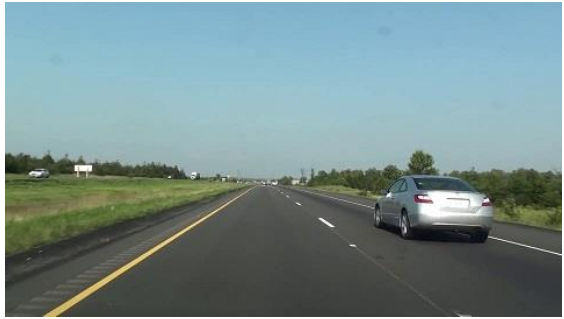 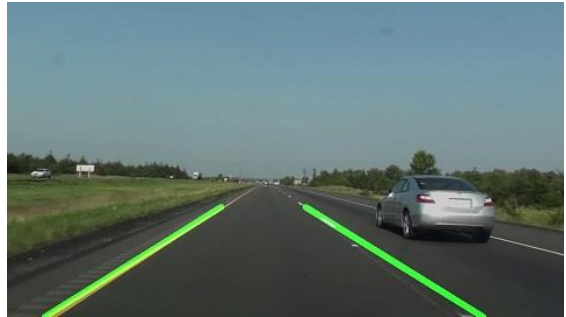

(c)

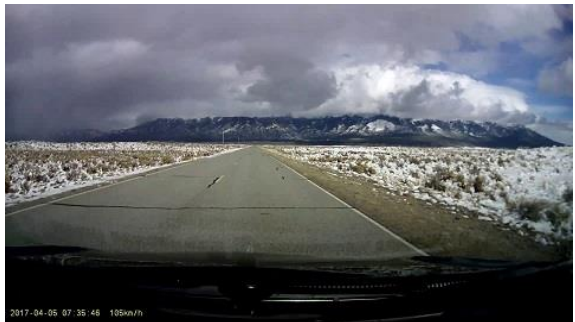 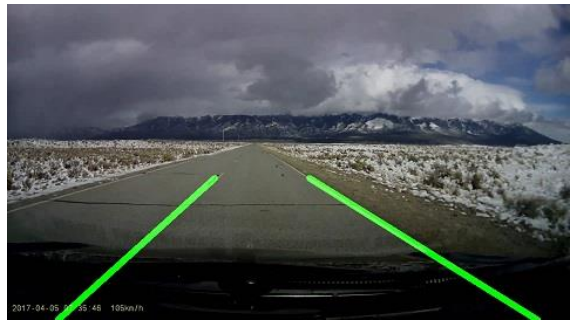

(d)

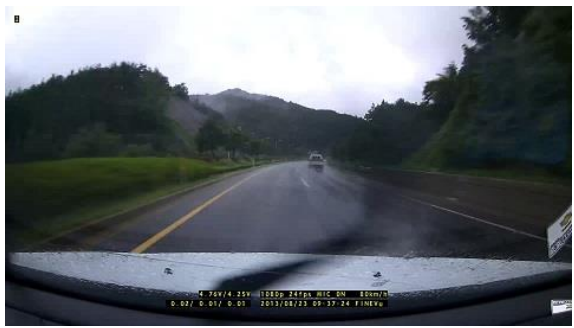 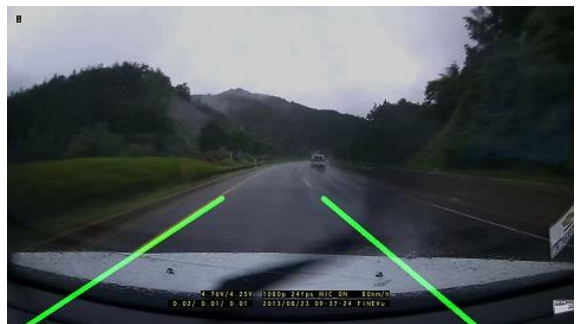

(e)

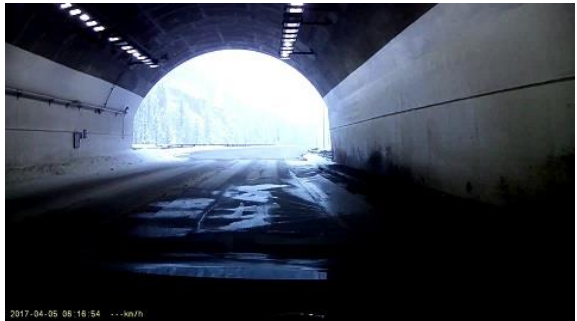 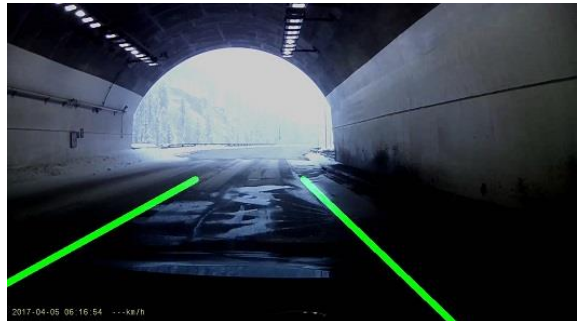

(f)

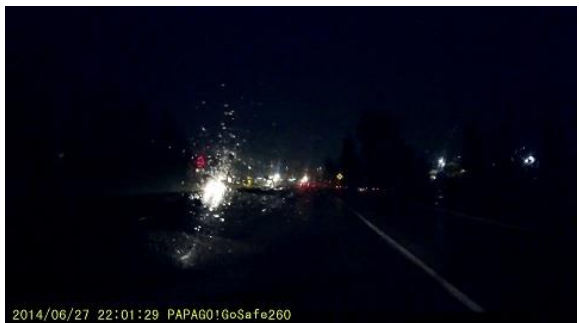 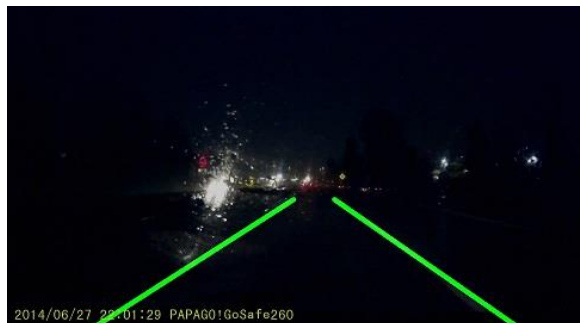

(g)

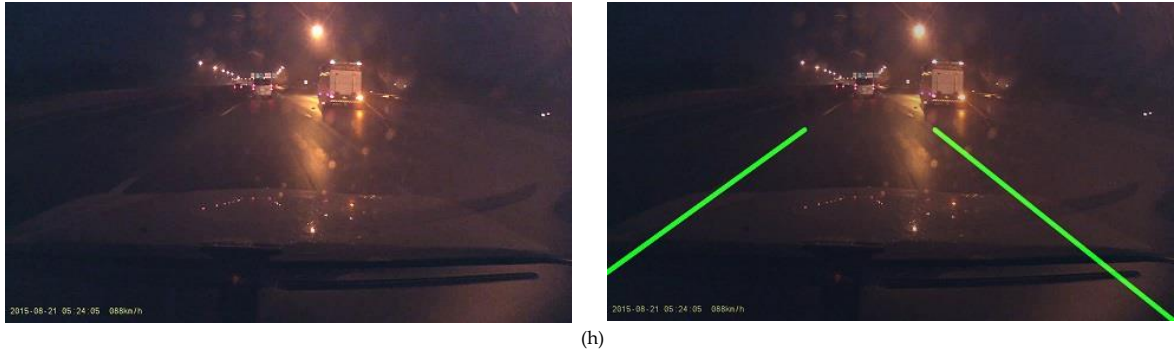
(h)

FIGURE 10. Successful lane detection and tracking under different challenges from DSDLDE dataset; (a) both lanes are eroded and right lane is occluded by wiper, (b) unpainted road, (c) right lane is dashed and slightly eroded, (d) left lane marking is cracked and fully eroded (e) right lane is dashed, blurred due to rain and occluded by wiper (f) both lane is almost eroded and fully invisible due to white out tunnel exit (g) left lane is invisible due to darkness and overexposed headlight reflection on windshield (h) right lane is invisible due to shiny effect created by street lamp and taillight reflection on wet windshield

## IV. Results

We performed experiments using the Spyder (python 3.7) environment on an Intel core i5, 2.30GHz CPU equipped with 8 GB RAM. We implemented the proposed method on two publicly available dataset: DSDLDE [1] with 1080x1920 resolutions at 24 frames/sec and SLD [2] with 480×720 resolutions at 25 frames/sec. DSDLDE, dataset is captured in USA and Korea. Recently, this dataset is being used to evaluate the performance of lane marking detection algorithms [15, 43] under harsh driving conditions e.g., adverse weather and critical lighting conditions. These videos are very challenging in itself as these are captured in different variations of weather conditions such as snow and rain. Moreover, it is captured in various illumination conditions such as abrupt light change while tunnel entry and exit, illumination variation captured in day and night. Experiments performed on 30 video clips which consist of more than 33 thousand frames. We also performed experiments on more than two thousand frames of SLD dataset including different environmental conditions like urban road, highway, lane change etc.

Two evaluation metrics are used to evaluate the performance of proposed method: the detection rate (%) and the processing time (ms/frame). As proposed method includes a lane tracking system, it can detect lane of each frame. Therefore, there is no such frame where lane is missed. As a result, after implementing proposed method, two types of frames we get as output: correctly detected frames and incorrectly detected frames. The detection rate is the number of correctly detected frames divided by the total number of frames. We determine that the detected frames are correct if the estimated line overlaid at least 70% of the actual lane line, the frame is considered as a correctly detected frame. The efficiency of our method is evaluated by detection rate (DR) which is calculated as follows:

$$DR = \frac{CDF}{TF} \times 100\%$$

Where DR, CDF and TF denote the detection rate, number of correctly detected frames, and number of total frames. Time complexity or processing time is the time that take to process each frame per second or millisecond. Every video has its own frame rate which define how many frames will be appeared within a second. A system is considered as real time if it can respond to events within specific time constraints. In general, standard video frame rate is 24 frame per second(fps). Therefore, the maximum processing time to be real time is:

$$T_m = \frac{1}{24}s = 0.04167s$$

The lane detection and tracking system should be less than 0.04167 s or 41.67 ms to be real time. Table 1 shows, the detection rate and processing time of proposed method performed on DSDLDE dataset under different time, weather, and challenging conditions. For DSDLDE dataset, the average detection rate is 97.36% and the average processing time is 29.06 msec per frame which meets the real time requirement. Our proposed method has solved lots of highly harsh driving conditions and abrupt illumination change at daytime i.e. darkness due to shadow, tunnel entry, extreme sunlight, white out on tunnel exit etc., abrupt illumination change at nighttime i.e. flare of light from other vehicle's headlight or taillight, reflection of road light on wet road, tunnel entry, tunnel exit etc., low contrast between lane marking and road surface when the lane is colored or eroded or the color of road light or tunnel light is same as lane color, incorrect lane detection due to presence of lines similar to lane i.e. guardrails, pavement marking, road divider, lines created by road crack, vehicle lines etc., detection interruption due to lane change, blurred view due to presence of heavy rain or snow, occlusion due to presence of wiper, raindrop, reflection of any stuff kept inside the car i.e. mobile, navigation device, video or image capturing device on the windshield. All these successful detections verify the robustness of the proposed algorithm.

TABLE 1. DETECTION RATES OF PROPOSED METHOD PERFORMED ON DSDLDE [1] DATASET

| Time | Weather | Condition/Challenges | number of frames($f_t$) | incorrectly detected frames($f_i$) | Detection rate, $(f_t - f_i)/f_t$ (%) | Average detection rate (%) | Processing time (ms/frame) |
|---|---|---|---|---|---|---|---|
| Day | Clear | Shadow, eroded lane marking, bumpy road | 4187 | 72 | 97.77 | 98.3 | 25.10 |
| | | Tunnel | 700 | 0 | 100 | | |
| | | Cracked road | 1122 | 33 | 97.06 | | |
| | Rainy | Light rain | 1463 | 20 | 98.63 | 96.9 | 30.20 |
| | | Heavy rain with traffic and lane change | 2522 | 130 | 94.85 | | |
| | | wet road inside tunnel | 500 | 16 | 96.80 | | |
| | | Hilly and curved highway | 1351 | 18 | 98.67 | | |
| | Snowy | Light snow, straight road | 2308 | 16 | 99.30 | 95.3 | 30.65 |
| | | Heavy snow, eroded lane marking, curved road, tunnel | 2092 | 191 | 90.86 | | |
| Night | Clear | Highway | 2455 | 10 | 99.51 | 99 | 25.5 |
| | | White tunnel | 900 | 8 | 99.11 | | |
| | | Yellow tunnel | 2100 | 35 | 97.61 | | |
| | Rainy | Light rain and clearly visible lane marking | 1463 | 0 | 100 | 96.5 | 31.39 |
| | | Light rain and traffic | 1235 | 6 | 99.51 | | |
| | | faded and unpainted lane marking, reflection of street-lamp on wet road | 1260 | 35 | 97.22 | | |
| | | Flare of headlight and taillight of car from same and opposite direction | 548 | 16 | 97.08 | | |
| | | Extreme dark (no road lamp) and bumpy road, reflection of taillight, lane change | 2680 | 100 | 96.26 | | |
| | | Reflection of navigation device and reflection of shiny bridge light on windshield, white tunnel | 1801 | 160 | 91.11 | | |
| | Snowy | Dark and curvy road | 1464 | 85 | 94.2 | 94.2 | 31.4 |
| Total and average | | | 33323 | 880 | 97.36% | | 29.06 |

TABLE 2. DETECTION RATE OF PROPOSED METHOD PERFORMED ON SLD [2] DATASET

| CHALLENGES | NUMBER OF FRAMES | DETECTION RATE (%) | PROCESSING TIME(MS/FRAME) |
|---|---|---|---|
| DAY URBAN | 998 | 97.90 | 15.55 |
| HIGHWAY AND LANE CHANGE | 1023 | 97.57 | 15.65 |
| TOTAL AND AVERAGE | 2021 | 97.73 | 15.60 |

Therefore, in the clear daytime the average lane detection rate is 98.07% which includes different conditions like shadow, eroded lane marking, lane occluded by traffic and pavement marking, cracked road, tunnel etc. In the rainy day, the average lane detection rate is 96.85% which includes conditions like heavy rain with traffic jam, lane change, wet road inside tunnel, hilly and curved highway. In the snowy day, the average lane detection rate is 95.29% which includes different difficult conditions like eroded lane marking, curved road, tunnel, wet road, blur view due to snow drop etc. In the clear night, the average detection rate is 98.3% which includes different conditions like city road with traffic, white and yellow colored light tunnel etc. In the rainy night, the average detection rate is 96.47% which includes some hard conditions like, faded and unpainted lane marking, reflection of road lamp on wet road, flare of headlight and taillight of car from same and opposite direction, Extreme dark (no road lamp) and bumpy road, reflection of taillight, lane change, Reflection of navigation device and reflection of shiny bridge light on windshield, white tunnel etc. In snowy night, the average detection rate is 94.19% where the visibility of lane is very poor due to snow fall at night. The overall detection rate in different weather conditions is more than 94%. Therefore, for DSDLDE dataset, the average detection rate is 97.36% and the average processing time is 29.06 msec per frame which meets the real time requirement.

Table 2 shows the detection rate of proposed method performed on SLD dataset. The detection rate is 97.90% in urban road condition and the detection rate is 97.57% in the highway and lane change scene. The average lane detection rate is 97.73% and the average processing time is 16.60 ms per frame which also meet the real time requirement. As the resolution of SLD dataset is lower than DSDLDE the processing time is less. Overall average detection rate of our proposed method is 97.55% and the average processing time is 22.32 msec per frame. The proposed method is compared with other methods that have been reported recently and have shown relatively better performance. In table 3, the performance of the proposed algorithm is compared with that of the previous works [1, 6, 9, 13] in terms of detection rate and processing time per frame in different weather and lighting conditions. The values in Table 3 are copied or estimated using the data of the reference papers mentioned above, although the video datasets are different. However, [1] and the proposed method is evaluated by the same dataset.

Our method outperforms son's [9] and jung's [13] method in terms of detection rate (%) and processing time(ms/frame) under all of the challenging conditions shown in table 3. In different challenging conditions e.g., snowy day, clear night, rainy night, and snowy night etc. the detection rates of proposed method are 2.3%, 1.03%, 2.9% and 2% higher than Lee's [1] detection rates respectively. However, the detection rate of Lee's method in clear day time is 0.7% higher than proposed method. It is important to mention that Lee does not bring up the number of frames used to evaluate performance under different conditions. The more frames used to evaluate a method, the more possibility of robustness as well as misdetection of that corresponding method. The detection rate of Marzougui [6] for night tunnel 0.51% higher than proposed method. The number of frames used to evaluate the performance of proposed method in night tunnel is twice the number of frames used to evaluate Marzougui's method in night tunnel. In addition, the number of frames used to evaluate proposed method in every weather and challenging conditions is higher than all other compared methods who mention the number of frames.

In terms of processing time, proposed method outperforms (10.68~16.68 ms faster) than state of the art methods [1, 9, 13]. However, Marzougui's [6] processing time is 0.79ms faster than proposed method because they performs experiments on low resolution images using sophisticated processing environment e.g., 2 cores higher processor clock speed.

TABLE 3. OVERALL PERFORMANCE COMPARISON OF PROPOSED METHOD IN DIFFERENT CONDITIONS

| | Detection rate (%) | | | | | | | | Processing time (ms/frame) |
|---|---|---|---|---|---|---|---|---|---|
| | Day | | | | Night | | | | |
| | Clear | rain | Tunnel | Snow | Clear | Rain | Tunnel | Snow | |
| Son [9] | 93.1(4656)* | 89 (1300)* | 93.2 (1000)* | - | 94.3 (3000)* | 93 (450)* | 98.1(790)* | - | 33 |
| Jung[13] | 88~98 | - | - | - | 87.7 | - | - | - | 39 |
| Lee[1] | **99** | 96.9 | - | 93 | 98 | 93.6 | - | 92.2 | 35.4 |
| Marzougui [6] | 90~93.5 (2218)* | 90.20 (847)* | - | - | 96.4(1440)* | - | 98.85 (1489)* | - | 21.54 |
| Proposed | 98.3 (6009)* | **96.9 (5836)*** | **98.4 (1200)*** | **95.3 (4400)*** | **99.03 (5455)*** | **96.5 (8987)*** | 98.34 (3000)* | **94.2 (1464)*** | 22.32 |

*Number of frames used to evaluate the performance

TABLE 4. PERFORMANCE COMPARISON OF THE PROPOSED METHOD ON SLD DATASETS

| METHODS | CONDITIONS | DETECTION RATE (%) | PROCESSING TIME(MS/FRAME) |
|---|---|---|---|
| Son [9] (1437 frames) | DAY HIGHWAY | 94.2 | 36.4 |
| Marzougui [6] (1136 frames) | DAY HIGHWAY | 92.7 | 21.54 |
| PROPOSED METHOD (2021 frames) | DAY HIGHWAY AND URBAN ROAD | 97.72 | 15.60 |

On top of everything else, proposed method outperforms Marzougui's method in terms of detection rate and processing time (summarized in table 4) when the dataset (SLD) and image resolution both are same to evaluate performance. The detection rate is 5.02% higher and processing time is 5.94ms faster than Marzougui's method. Table 3 and 4 shows that the proposed method yields better result than all other methods and more challenges related to light, road and weather conditions are explored in this work than the other state of the art works. Especially, video clips of roads at snowy night, did not get good result in any of the other works. The robustness is the key that determine if a system is appropriate for applying in real life. From the above comparison it is proved that proposed method is robust and appropriate for real-life application. However, the proposed algorithm failed to detect lane markings when snow creates lane like lines and when tunnel edges create lane like lines as shown in Figure 11. There exist some cases of misdetection of lane marking. It could be improved by adding more parameters in lane verification stage.

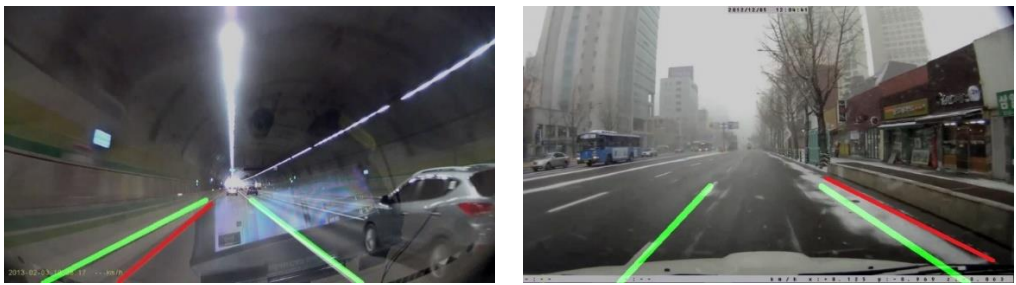

FIGURE 11. Example of failed lane marking detection, red line indicates missed lane marking

## V. Conclusion

In this paper, we have addressed all kind of real-life environmental challenges to detect road lane markings and categorized them into four types: first, abrupt illumination changes due to change of time, weather, road etc., second, lane markings get obscured partially or fully when the lane markings are colored, eroded or occluded, third, blurred view created by adverse weather(rain/snow), fourth, incorrect lane detection due to presence of lane like confusing lines. We have proposed a robust method to detect road lane marking under all these challenging real-life environmental conditions. A comprehensive intensity threshold range (CITR) is proposed in edge detection stage, which improves the performance of canny. A two-step lane verification technique, angle-based geometric constraint (AGC) and length based geometric constraint (LGC) algorithm is proposed, to verify the characteristics of lane marking. Finally, a novel lane tracking method, which predict the lane position of next frame by defining a range of horizontal lane position (RHLP) of next frame with respect to horizontal lane position of previous frame is introduced to keep track of the lane position when either left or right or both lane markings are partially or fully invisible due to erosion or occlusion for short period of time. The proposed algorithm shows better performance for road conditions with noisy components than others. The proposed algorithm is verified by 30 video clips consist of various challenges. The computation time satisfies the real-time operation. The detection rate and the computation time for the proposed method are compared with those of other works and it is manifested that the proposed method is superior to them.

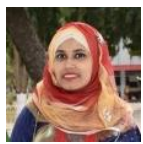
SAMIA SULTANA received the B.Sc. degree in Computer Science and Engineering in 2013, and the M.Sc. degree in Computer Science and Engineering in 2021 from the Rajshahi University of Engineering & Technology, Bangladesh. She is working as a Sessional Lecturer in School of IT and Engineering, Melbourne Institute of Technology, Sydney Campus, Australia.

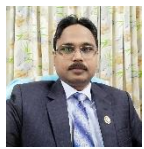
BOSHIR AHMED is working as a faculty member in the department of Computer Science & Engineering, Rajshahi University of Engineering & Technology since 24th June 2001. In addition to this role, he is professor of this Department, with research interests in image/video analysis and Computer vision. In this service period, he took active part in postgraduate & undergraduate academic programs of this department and have got the opportunity to conduct and supervise different research projects.

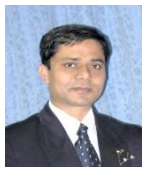
SHAMIM AHMAD is a professor in the department of Computer Science and Engineering at Rajshahi University. He received the Ph.D. degree in 2005 from Chubu University, Japan. He received the M.Sc. degree in Computer Science and Engineering in 1992, and the B.Sc. degree in Computer Science and Engineering in 1991 from the Rajshahi University, Bangladesh.

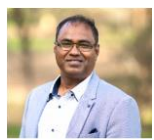
MANORANJAN PAUL (Senior Member, IEEE) received the Ph.D. degree from Monash University, Australia, in 2005. He was a Postdoctoral Research Fellow with the University of New South Wales, Monash University, and Nanyang Technological University. He is currently a Full Professor, the Director of the Computer Vision Laboratory, and the Leader of the Machine Vision & Digital Health (MaViDH) Research Group, Charles Sturt University, Australia. He has published around 200 peer reviewed publications, including 72 journals. His research interests include video coding, image processing, digital health, wine technology, machine learning, EEG signal processing, eye tracking, and computer vision. He was an Invited Keynote Speaker in IEEE DICTA-17 & 13, CWCN-17, WoWMoM-14, and ICCIT-10. He is also an Associate Editor of three top ranked journals, such as IEEE TRANSACTIONS ON MULTIMEDIA, IEEE TRANSACTIONS ON CIRCUITS AND SYSTEMS FOR VIDEO TECHNOLOGY, and *EURASIP Journal in Advances on Signal Processing*. He was the General Chair of PSIVT-19 and the Program Chair of PSIVT-17 and DICTA-18. He was awarded the ICT Researcher of the Year 2017 by Australian Computer Society. He obtained more than $3.6 million competitive external grant, including Australian Research Council (ARC) Discovery Grants and Australia China Grant.

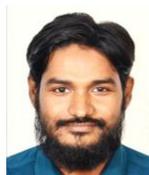
MUHAMMAD RAFIQUL ISLAM received the B.Sc. degree in electrical and electronic engineering from the Rajshahi University of Engineering & Technology, Bangladesh, in 2011, and the M.Sc. degree in electrical, electronic and system engineering from The National University of Malaysia, in 2016. He is currently pursuing the Ph.D. degree with the School of Computing and Mathematics, Charles Sturt University, Australia. From 2013 to 2016, he was a Graduate Research Assistant with three different research grants at The National University of Malaysia. Mr. Islam's awards and honours include the AGRTP-International (Australian Government Research Training Program) Scholarship, the Best Paper Award in HDR conference, in 2019, Charles Sturt University. He is currently working as a Sessional Lecturer in School of IT and Engineering, Melbourne Institute of Technology, Sydney Campus, Australia.